\documentclass{article}
\usepackage[utf8]{inputenc}
\usepackage[T1]{fontenc}
\usepackage{authblk}
\usepackage{booktabs}
\usepackage{graphicx}
\usepackage{tabularx}
\usepackage{array}
\usepackage{multirow}
\usepackage[margin=0.75in]{geometry}
\usepackage{xurl}
\usepackage{microtype}
\usepackage{placeins}
\newcommand{\keywords}[1]{\noindent\textbf{Keywords: } #1}

\newcommand{\file}[1]{\path{#1}}
\newcolumntype{Y}{>{\raggedright\arraybackslash}X}
\title{A Multi-Center Breast FNAC Whole-Slide Cytology Dataset for AI-Assisted Patch-Wise Classification Using C1–C5 Reporting Categories}
\date{}
\author[1]{Garima Jain}
\author[2]{Abhijeet Patil}
\author[3]{Surabhi Jain}
\author[4]{Sanghamitra Pati}
\author[2,*]{Amit Sethi}
\author[3]{Sandeep Mathur}
\author[1]{Pulkit Verma}
\author[1]{Nishi Halduniya}
\author[1]{Jatin Kashyap}
\author[1]{Sharat Kumar}
\author[6]{Simmi Kharb}
\author[7]{Sunita Singh}
\author[8]{Sucheta Devi Khuraijam}
\author[8]{Sushma Khuraijam}
\author[8]{Ratan Konjengbam}
\author[9]{Arvind Kumar}
\author[9]{Deepali Tirkey}
\author[9]{Saurav Banerjee}
\author[10]{Shivani Kalhan}
\author[10]{Rakesh Kumar Gupta}
\author[11]{Ranjana Solanki}
\author[11]{Deepika Hemranjani}
\author[12]{Shashank Nath Singh}
\author[13]{Uma Handa}
\author[13]{Manveen Kaur}
\author[14]{B. G. Malathi}
\author[14]{Yogender P.}
\author[15]{Niraj Kumari}
\author[15]{Shruti Gupta}
\author[16]{Indu R. Nair}
\author[16]{Vidya C.}
\author[17]{Basumitra Das}
\author[17]{Sunil Kumar Komanapalli}
\author[18]{Ravindra Karle}
\author[18]{Tanaya Kulkarni}
\author[19]{Vandana Raphael}
\author[19]{Biswajit Dey}
\author[5]{Vaishali Gaikwad}
\author[5]{Nilam Adhav}
\affil[1]{Indian Council of Medical Research-National Institute for Research in Digital Health \& Data Science, New Delhi}
\affil[2]{Department of Electrical Engineering, Indian Institute of Technology Bombay, Mumbai}
\affil[3]{Department of Pathology, All India Institute of Medical Sciences, New Delhi}
\affil[4]{Indian Council of Medical Research, New Delhi}
\affil[5]{Department of Pathology, Lokmanya Tilak Municipal Medical College and General Hospital, Sion, Mumbai}
\affil[6]{Department of Biochemistry, Pt. B.D. Sharma Post Graduate Institute of Medical Sciences, Rohtak}
\affil[7]{Department of Pathology, Pt. B.D. Sharma Post Graduate Institute of Medical Sciences, Rohtak}
\affil[8]{Department of Pathology, Regional Institute of Medical Sciences, Imphal}
\affil[9]{Department of Pathology, Rajendra Institute of Medical Sciences, Ranchi}
\affil[10]{Government Institute of Medical Sciences, Greater Noida}
\affil[11]{Department of Pathology, Sawai Man Singh Medical College, Jaipur}
\affil[12]{Department of Otorhinolaryngology, Sawai Man Singh Medical College, Jaipur}
\affil[13]{Department of Pathology, Government Medical College and Hospital, Chandigarh}
\affil[14]{Department of Pathology, Mandya Institute of Medical Sciences, Mandya}
\affil[15]{Department of Pathology and Lab Medicine, All India Institute of Medical Sciences, Raebareli}
\affil[16]{Department of Pathology, Amrita Institute of Medical Sciences, Kochi}
\affil[17]{Department of Pathology, Government Medical College, Vizianagaram}
\affil[18]{Department of Pathology, Dr. Balasaheb Vikhe Patil Rural Medical College, Loni}
\affil[19]{Department of Pathology, North Eastern Indira Gandhi Regional Institute of Health and Medical Sciences, Shillong}
\affil[*]{Correspondence: asethi@iitb.ac.in}

\begin{document}
\maketitle

\begin{abstract}
We present a multi-center breast fine needle aspiration cytology (FNAC) dataset designed for patch-wise classification using C1--C5 reporting labels. The prospective dataset includes 321 patients and 470 whole-slide images (WSIs) collected from participating tertiary medical centers in India between May 2023 and March 2026. Slides were stained using Papanicolaou (190 WSIs) or May--Gr\"unwald--Giemsa (280 WSIs), scanned on a Hamamatsu NanoZoomer S360 at 40$\times$ magnification and 0.25 microns per pixel, and stored directly in NDPI format. Across the 470 WSIs, 446 WSIs contain annotated patch regions, yielding 7,398 PNG image patches with expert-verified C1--C5 labels. The release includes NDPI WSIs, WSI-level GeoJSON annotation files, extracted patch images, de-identified metadata, a data dictionary, a validation summary, a manifest linking WSIs to Zenodo records, and code for dataset inspection and reuse. The complete dataset is approximately 950 GB and is available through Zenodo.
\end{abstract}

\flushbottom
\thispagestyle{empty}
\section*{Background and Summary}

Breast cancer is one of the most common malignancies worldwide and remains a major cause of cancer-related morbidity and mortality among women \cite{Bray:2024ca,Sathishkumar:2022ijmr}. Early and accurate diagnosis of breast lesions is essential for timely clinical management. In many clinical settings, especially in resource-constrained regions, fine needle aspiration cytology (FNAC) remains an important method for the initial evaluation of palpable and radiologically detected breast lesions. FNAC is minimally invasive, rapid, cost-effective, and clinically useful when performed and interpreted by trained personnel \cite{Mendoza:2011pri,Yu:2012bmc,Willems:2012jcp,Wang:2017breast}.

Breast FNAC is commonly used alongside clinical examination and radiological assessment as part of a triple-assessment approach \cite{Morris:1998archsurg,Morris:2001archsurg}. Cytological interpretation requires assessment of cellularity, architectural arrangement, nuclear features, cytoplasmic characteristics, background elements, and features suggestive of benignity, atypia, suspicion, or malignancy. Interpretation can be challenging because benign proliferative lesions, atypical lesions, low-grade malignancies, and poorly preserved or scanty samples may show overlapping cytomorphological features \cite{Sidawy:1998diagncyto,Mitra:2015jcytol,Kanhoush:2004cancer}. To improve standardization, the International Academy of Cytology (IAC) Yokohama System defines five breast FNAC reporting categories: C1 insufficient or inadequate, C2 benign, C3 atypical, C4 suspicious for malignancy, and C5 malignant \cite{Field:2017actacytol,Field:2019actacytol}. These categories are associated with diagnostic criteria, risk of malignancy, and suggested clinical management \cite{Field:2019actacytol,Hoda:2019actacytol,Paul:2023actacytol,Nikas:2023ajcp}.

Digital cytology and artificial intelligence (AI)-assisted analysis have the potential to support region triaging, patch-level classification, benign-versus-malignant classification, and future slide-level decision-support workflows \cite{Niazi:2019lancet,Dimitriou:2019frontmed,Jiang:2023mia,Thakur:2022cancers}. However, development of robust models depends on access to large, diverse, well-annotated, and technically validated datasets. In cytopathology, such datasets are difficult to construct because they require expert review, careful quality control, de-identification, and coordination across medical centers \cite{Jiang:2023mia,Thakur:2022cancers}. Recent multi-center cytology dataset efforts in oral cytology have highlighted the value of releasing whole-slide images, annotations, metadata, and reusable code for AI development \cite{Jain:2025oralcyto}. For breast cytology, established resources such as the Wisconsin Diagnostic Breast Cancer dataset and BreakHis are based on extracted nuclear features or histopathology images rather than multi-center breast FNAC WSIs with C1--C5 patch-level cytology labels \cite{Street:1993spie,Spanhol:2016tbme}. Prior breast FNAC image studies have generally used smaller cytology image datasets and binary benign-versus-malignant classification, limiting reproducible benchmarking of C1--C5 breast cytology models \cite{Saikia:2019tice,Zerouaoui:2022healthinf}.

To address this gap, we release a multi-center breast FNAC cytology dataset for patch-wise classification using C1--C5 reporting labels. The dataset includes 321 patients and 470 digitized WSIs stained using Papanicolaou (PAP) and May--Gr\"unwald--Giemsa (MGG) protocols. It contains 7,398 expert-verified image patches extracted from diagnostically relevant regions. The category-wise patch distribution is 33 C1 patches, 3,706 C2 patches, 478 C3 patches, 402 C4 patches, and 2,779 C5 patches. The release also includes WSI-level GeoJSON annotation files, patch images, de-identified metadata files, a data dictionary, a validation summary, and a manifest linking WSI files to the corresponding Zenodo records. By providing both WSIs and extracted patches, the dataset supports direct patch-level experiments while retaining traceability from each patch to its parent WSI and spatial location.
\begin{table}[ht]
\centering
\small
\caption{Dataset key statistics}
\label{tab:stats}
\begin{tabularx}{\textwidth}{|p{0.22\textwidth}|Y|}
\hline
\textbf{Attribute} & \textbf{Key statistics} \\ \hline
Number of Patients & 321 \\ \hline
Number of WSIs & 470 \\ \hline
Stain types & PAP -- 190, MGG -- 280 \\ \hline
Age & Range -- 12 to 90 years\newline Mean -- 36 years\newline Median -- 34 years \\ \hline
Diagnosis & Non-cancerous -- 209 (304)\newline Cancerous -- 112 (166) \\ \hline
Category & C1 insufficient -- 4 (9)\newline C2 benign -- 194 (281)\newline C3 atypical -- 11 (14)\newline C4 suspicious for malignancy -- 11 (15)\newline C5 malignant -- 101 (151) \\ \hline
Participating centers & Pandit Bhagwat Dayal Sharma Post Graduate Institute of Medical Sciences, Rohtak -- 65 (73)\newline Regional Institute of Medical Sciences, Imphal -- 60 (88)\newline Rajendra Institute of Medical Sciences, Ranchi -- 43 (72)\newline Government Institute of Medical Sciences, Noida -- 35 (53)\newline Sawai Man Singh Medical College, Jaipur -- 35 (38)\newline Government Medical College and Hospital, Chandigarh -- 28 (41)\newline Mandya Institute of Medical Sciences, Mandya -- 10 (28)\newline AIIMS, Raebareli -- 12 (12)\newline Amrita Hospital, Kochi -- 7 (10)\newline Government Medical College, Vizianagaram (GMCV) -- 6 (11)\newline Dr. Balasaheb Vikhe Patil RMC, Loni -- 6 (18)\newline NEIGRIHMS, Meghalaya -- 6 (18)\newline Lokmanya Tilak MMC, Mumbai -- 8 (8) \\ \hline
\end{tabularx}
\end{table}

\section*{Methods}
\begin{figure}[htbp]
    \centering
    \includegraphics[width=0.75\linewidth]{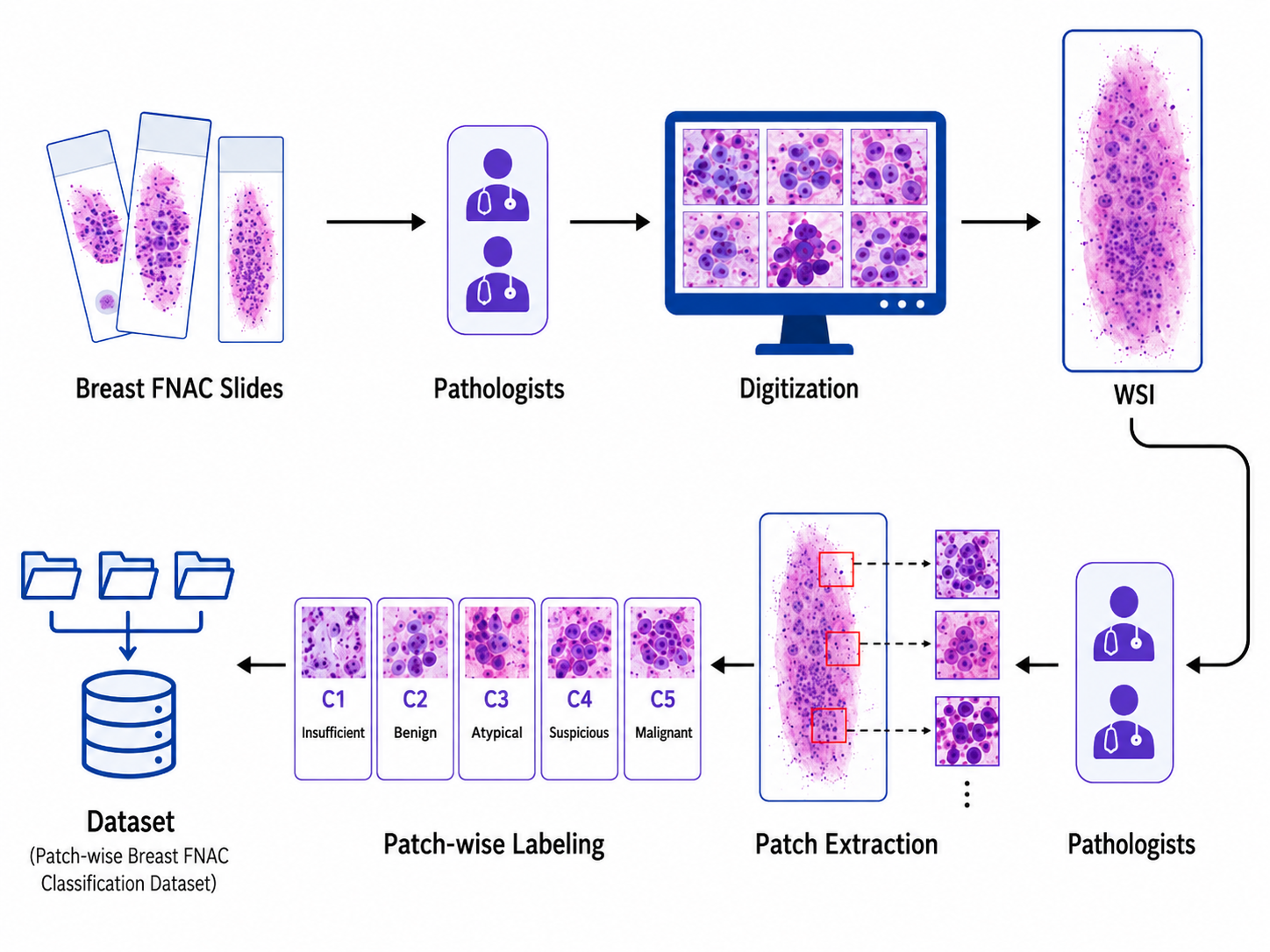}
    \caption{Overall workflow for breast FNAC dataset creation, slide digitization, WSI-level patch annotation, patch extraction, patch-wise labeling, and dataset release.}
    \label{fig:dataset}
\end{figure}

We describe the sample collection, slide digitization, quality control, WSI-level annotation, patch extraction, and patch-wise labeling process used to construct the breast FNAC cytology dataset. A prospective multi-center cohort study was conducted across participating tertiary medical centers in India. Ethical clearance was obtained from all participating centers and the Indian Council of Medical Research (ICMR). The study was carried out from May 2023 to March 2026. The dataset includes breast FNAC slides from 321 patients and 470 whole-slide images (WSIs), stained using Papanicolaou (PAP) and May--Gr\"unwald--Giemsa (MGG) protocols. The summary of the dataset is shown in Table~\ref{tab:stats}. The dataset was designed for supervised patch-wise classification according to the five breast FNAC reporting categories C1 to C5. The overall workflow for dataset preparation is shown in Figure~\ref{fig:dataset}.

\subsection*{Sample collection and slide quality control}
Breast FNAC samples were collected at participating medical centers as part of routine cytopathological evaluation. FNAC was performed using the standard procedure generally followed in cytology practice. Aspirated cellular material was smeared onto clean glass slides for cytological examination. Relevant patient-level and slide-level clinical details were collected using cytopathology requisition forms. A subset of slides was fixed and stained using the PAP stain, while the remaining slides were air dried and stained using the MGG stain. The final dataset includes 190 PAP-stained WSIs and 280 MGG-stained WSIs. Slides with missing identifiers, improper labeling, incomplete documentation, broken slides, poor staining, low cellularity, poor focus, or scanning artifact were excluded from downstream dataset preparation. WSI-level quality control was performed manually by the pathology and data teams. Slides with focus issues or scanning artifacts were rescanned where applicable. All cytology slides were reviewed by pathology experts. In total, 209 patients corresponding to 304 WSIs were categorized as non-cancerous, while 112 patients corresponding to 166 WSIs were categorized as cancerous.

\subsection*{Digitization}
All slides were digitized using the same Hamamatsu NanoZoomer S360 whole-slide scanner at 40$\times$ magnification. WSIs were acquired at a spatial resolution of 0.25 microns per pixel and saved directly in NDPI format. The digitization workflow generated 470 WSIs from 321 patients. Since slides were collected from multiple participating centers, the dataset captures variability in patient population, sample preparation, staining characteristics, cellularity, and cytomorphological appearance, while scanner-related variability is limited by the use of a single scanner model and a standardized scanning workflow.

\subsection*{WSI-level annotation and patch extraction}
Diagnostically relevant regions were manually identified on the digitized breast FNAC WSIs. WSI-level rectangular box annotations were created in QuPath and exported as GeoJSON files. Annotation coordinates are stored in the level-0 WSI pixel coordinate system. Each GeoJSON annotation is a rectangular box, and the patch label is stored in the GeoJSON \file{classification.name} property as one of the Roman numeral labels \file{I}, \file{II}, \file{III}, \file{IV}, or \file{V}. These labels map to C1 insufficient, C2 benign, C3 atypical, C4 suspicious for malignancy, and C5 malignant, respectively.

A GeoJSON file is provided for each of the 470 WSIs. Of these, 446 WSIs contain one or more annotated patch regions, while 24 WSIs have no extracted patch regions. Extracted patches correspond exactly to the annotated bounding-box regions and were saved as PNG files. Patch filenames encode the parent WSI and level-0 coordinate of the extracted region using the format \file{wsi-name_x-coord_y-coord.png}. Patch dimensions are variable because the manually selected rectangular annotations vary in size according to the diagnostic region and cytological content. Across the 446 annotated WSIs, 7,398 image patches were extracted. The category-wise distribution of extracted patches is summarized in Table~\ref{tab:dataset_summary}.

\subsection*{Patch-wise labeling}
Each extracted patch was assigned one of five breast FNAC reporting labels: C1, C2, C3, C4, or C5. Patch labels were assigned independently at the patch level while reviewing the corresponding WSI; they were not simply inherited from the slide-level category. A pathologist initially selected and labeled diagnostically relevant WSI regions, and a senior pathologist subsequently verified the patch labels. Corrections were made after discussion between the pathologist and senior pathologist wherever required. Only the final verified labels are retained in the released dataset. The final labels should therefore be interpreted as expert-verified reference labels for patch-level cytomorphological classification. Since the dataset is patch-based, the labels correspond to the diagnostic interpretation of individual image patches and are not intended to replace final patient-level diagnosis. Final clinical diagnosis may require integration of slide-level findings, clinical history, radiological findings, and histopathological correlation. The final dataset contains 33 patches from C1, 3,706 patches from C2, 478 patches from C3, 402 patches from C4, and 2,779 patches from C5, resulting in a total of 7,398 labeled patches.

\begin{table}[hb]
\centering
\small
\caption{Number of extracted patches across five breast FNAC reporting categories (C1--C5).}
\label{tab:dataset_summary}
\begin{tabular}{lcccccc}
\toprule
 & \textbf{C1} & \textbf{C2} & \textbf{C3} & \textbf{C4} & \textbf{C5} & \textbf{Total} \\
\midrule
\# of patches  & 33  & 3706 & 478 & 402 & 2779 & 7398 \\
\bottomrule
\end{tabular}
\end{table}

\FloatBarrier
\section*{Ethics Statement}
The study protocol was reviewed and approved by the Institutional Ethics Committees/Institutional Review Boards of the participating medical centers and the Indian Council of Medical Research. Ethics approval was obtained from the Institutional Ethics Committee of ICMR--National Institute for Research in Digital Health \& Data Science, New Delhi, approval no. ICMR-NIP-IEC/29-12-2021/05/02R1; Regional Institute of Medical Sciences, Imphal, approval no. A/2026/REB/Prop(FP)190/118/29/2022; Pt. B.D. Sharma Post Graduate Institute of Medical Sciences, Rohtak, approval no. BREC/22/071; Rajendra Institute of Medical Sciences, Ranchi, approval no. ECR/769/INST/JH/2015/RR-21; Government Institute of Medical Sciences, Noida, approval no. GIMS/IEC/HR/2022/41; Government Medical College and Hospital, Chandigarh, approval no. GMCH/IEC/715R/2022/103; Sawai Man Singh Medical College, Jaipur, approval no. 568/MC/EC/2022; Mandya Institute of Medical Sciences, Mandya, approval no. MIMS/IEC/2024/969; Dr. Balasaheb Vikhe Patil Rural Medical College, Loni, approval no. PIMS/IEC-DR/2024/342; North Eastern Indira Gandhi Regional Institute of Health and Medical Sciences, Shillong, approval no. NEIGR/IEC/M14/F6/2024; Government Medical College, Vizianagaram, approval no. EC/NEW/INST/2023/3519; Amrita Institute of Medical Sciences, Kochi, approval no. IEC-AIMS-2024-PATHO-216; All India Institute of Medical Sciences, Raebareli, approval no. F-7/BIOETHICS/AIIMS-ABL/APPR/EM/2024-8/3; Lokmanya Tilak Municipal Medical College and General Hospital, Mumbai, approval no. IEC/770/24. The respective ethics committees approved the study protocol, participant recruitment, collection of clinical and cytology-related data, digitization of cytology slides/images, and sharing of de-identified study data/images among collaborating study investigators and institutions for research and analysis related to the approved protocol.

Written informed consent was obtained from all adult participants prior to enrolment in the study. For participants below 18 years of age, written informed consent was obtained from a parent or legal guardian, and assent was obtained from minor participants wherever applicable. The consent included permission for participation, collection of relevant clinical information and biological/cytology material, digitization of slides/images, and use and sharing of de-identified data/images for research, analysis, publication, and development or validation of digital pathology and artificial intelligence-based methods, in accordance with the approved study protocol and applicable ethical guidelines.

\section*{Data Records}
The breast FNAC cytology dataset is available at Zenodo \cite{Patil:2026zenodo}. The main Zenodo record contains \file{breast-cytology-main.zip}, which includes extracted patch images, WSI-level GeoJSON annotation files, public metadata files, and a manifest linking the WSI files to the corresponding Zenodo records. The complete dataset, including the 470 whole-slide images, is approximately 950 GB. The WSI files are provided in NDPI format and were generated directly from breast FNAC slides scanned using a Hamamatsu NanoZoomer S360 scanner at 40$\times$ magnification. The WSIs are distributed across 21 linked Zenodo records because of file-size limits; the linked WSI records contain 26, 19, 25, 22, 18, 17, 28, 26, 36, 24, 26, 21, 18, 20, 18, 18, 28, 20, 26, 28, and 6 WSIs, respectively, for a total of 470 WSIs.

The main record has the following top-level organization:
\begin{verbatim}
breast-cytology-main.zip
  geojson/
  patches/<WSI_ID>/<I-V>/<patch files>
  metadata/
  manifest/zenodo_files_manifest.csv
\end{verbatim}

The \file{geojson/} directory contains 470 WSI-level GeoJSON files, one for each WSI. GeoJSON files for the 446 annotated WSIs contain rectangular patch-coordinate annotations in the level-0 WSI coordinate system. The remaining 24 WSIs have GeoJSON files but no extracted patch regions. The annotation property \file{classification.name} stores the Roman numeral label \file{I}--\file{V}, corresponding to C1--C5. These WSI-level annotation files allow extracted patches to be traced back to their source WSI and spatial location.

The \file{patches/} directory contains PNG image patches extracted exactly from the annotated rectangular regions. Patches are organized by parent WSI identifier. Within each WSI directory, patches are grouped into category directories labelled \file{I} to \file{V}. These category directories correspond to the C1--C5 diagnostic labels as follows: \file{I}=C1 insufficient, \file{II}=C2 benign, \file{III}=C3 atypical, \file{IV}=C4 suspicious for malignancy, and \file{V}=C5 malignant. Patch filenames follow the format \file{wsi-name_x-coord_y-coord.png}, where the coordinates denote the level-0 WSI location of the extracted region. This structure provides direct access to diagnostically relevant image regions for patch-wise machine learning experiments without requiring users to process the full WSI files first.

The \file{metadata/} directory contains \file{master_metadata.csv}, \file{patient_metadata.csv}, \file{slide_metadata.csv}, \file{site_metadata.csv}, \file{class_summary.csv}, \file{dataset_summary.csv}, \file{metadata_validation_summary.csv}, \file{data_dictionary.csv}, and \file{README_metadata_release.md}. These files provide de-identified patient identifiers, slide identifiers, stain type, diagnostic group, C1--C5 category information, collection site, WSI-level and patch-level linkage information, class summaries, dataset-level summaries, metadata validation results, and definitions of released variables and allowed values. The \file{data_dictionary.csv} and \file{README_metadata_release.md} files are included to support interpretation and reuse of the released metadata.

The \file{manifest/} directory contains \file{zenodo_files_manifest.csv}. This manifest records the released WSI filenames, Zenodo record URLs, file sizes, and checksums for the distributed WSI records. The manifest is intended to help users verify file organization, locate WSIs across the distributed records, and link each WSI to its annotations, patches, and metadata entries.

\begin{table}[ht]
\centering
\small
\caption{Overview of the data records included in the breast FNAC cytology dataset.}
\label{tab:data_records}
\begin{tabularx}{\textwidth}{p{0.28\textwidth}Y}
\hline
\textbf{Data record} & \textbf{Description} \\
\hline
Whole slide images (WSIs) & 470 breast FNAC WSIs in NDPI format, scanned at 40$\times$ magnification using a Hamamatsu NanoZoomer S360 scanner and distributed across 21 linked Zenodo records listed in \file{zenodo_files_manifest.csv}. \\
WSI-level GeoJSON annotations & 470 GeoJSON files in \file{geojson/}; 446 contain rectangular patch-coordinate annotations and \file{classification.name} labels, while 24 correspond to WSIs without extracted patch regions. \\
Extracted patches & 7,398 PNG patch image files in \file{patches/}, organized by parent WSI and by category directories \file{I}--\file{V}, corresponding to C1--C5. \\
Metadata files & Public metadata CSV files in \file{metadata/}, including patient-, slide-, site-, class-summary, dataset-summary, master-metadata, validation-summary, and data-dictionary files, together with a metadata README. \\
Manifest & \file{zenodo_files_manifest.csv}, listing WSI file organization, Zenodo record URLs, file sizes, and checksums for the linked WSI records. \\
\hline
\end{tabularx}
\end{table}
\section*{Technical Validation}
Technical validation was performed to verify the completeness, consistency, readability, traceability, and reuse readiness of the released dataset. Validation checks were conducted across the WSI records, GeoJSON annotation files, extracted patch images, public metadata files, and dataset manifest. The 21 linked WSI records were checked against the released manifest to confirm that they collectively account for 470 NDPI WSIs. WSI filenames in the manifest were checked for consistency with the corresponding entries in the slide-level and master metadata files.

The 470 WSI-level GeoJSON files were checked for correspondence with the released WSI identifiers. GeoJSON files were parsed to confirm readability, coordinate validity, use of rectangular patch-coordinate annotations, and use of the expected \file{classification.name} property. Validation confirmed 446 WSIs with one or more annotated patch regions and 24 WSIs without extracted patch regions. The Roman numeral labels \file{I}--\file{V} were checked for consistency with the C1--C5 reporting categories recorded in the metadata and patch-directory organization.

Patch-level validation included checking the presence and readability of 7,398 PNG patch image files, confirming parent WSI linkage, checking patch filename coordinate fields, and verifying category-directory organization. Patch-level class assignments were checked for consistency with the WSI-level GeoJSON annotations and released metadata. Because patch regions were manually selected to preserve diagnostically meaningful cytological content, patches vary in image size. This size variability was retained in the released data and should be handled during model development according to the requirements of the downstream pipeline.

Metadata validation included checking consistency across \file{master_metadata.csv}, \file{patient_metadata.csv}, \file{slide_metadata.csv}, \file{site_metadata.csv}, \file{class_summary.csv}, and \file{dataset_summary.csv}. Validation checks confirmed the expected totals of 321 patients, 470 WSIs, 7,398 patches, 190 PAP-stained WSIs, 280 MGG-stained WSIs, and the reported C1--C5 patient-, WSI-, and patch-level counts. Patient identifiers, slide identifiers, stain type, diagnostic group, C1--C5 category, site information, patch availability fields, and WSI-to-file linkage fields were checked for internal consistency. The file \file{metadata_validation_summary.csv} summarizes the validation checks performed before release. De-identification checks were performed before public release to ensure that direct patient identifiers were removed from shared image files and metadata.

\begin{table}[ht]
\centering
\small
\caption{Dataset-level validation checks performed before release.}
\label{tab:technical_validation}
\begin{tabularx}{\textwidth}{p{0.28\textwidth}Y}
\hline
\textbf{Dataset component} & \textbf{Validation check} \\
\hline
WSI records & Confirmed that the 21 linked Zenodo WSI records collectively account for 470 NDPI WSIs and are represented in \file{zenodo_files_manifest.csv}. \\
GeoJSON annotations & Checked that 470 WSI-level GeoJSON files are present, parseable, and linked to the expected WSI identifiers; 446 WSIs contain patch annotations and 24 do not contain extracted patch regions. \\
Patch coordinates & Checked that rectangular patch-coordinate annotations use level-0 WSI coordinates and can be traced from GeoJSON files to extracted patch images and parent WSIs. \\
Patch images & Checked presence, readability, parent WSI linkage, coordinate-encoded filenames, and category-directory organization of 7,398 PNG patch files. \\
Class labels & Checked use of the expected \file{I}--\file{V} GeoJSON and folder labels and their mapping to C1--C5 reporting categories. \\
Metadata files & Checked consistency across master, patient-level, slide-level, site-level, class-summary, dataset-summary, and validation-summary metadata tables. \\
Data dictionary and README & Provided definitions for released variables, allowed values, file organization, and metadata interpretation. \\
De-identification & Confirmed removal of direct patient identifiers from public metadata and released image records. \\
\hline
\end{tabularx}
\end{table}
\section*{Usage Notes}
The released labels should be interpreted as expert-verified reference labels for patch-wise cytomorphological classification rather than final patient-level diagnoses. Final clinical interpretation of breast FNAC may require integration of slide-level cytology, clinical history, radiological findings, and histopathological correlation. Users developing slide-level or patient-level models should therefore aggregate patch-level information cautiously and should avoid interpreting a single patch label as a complete diagnostic conclusion.

The dataset is diagnostically enriched because patches were manually selected from interpretable and diagnostically relevant regions rather than sampled exhaustively from each WSI. The released patch annotations are not complete annotations of every cytological element on each slide. Of the 470 WSIs, 446 contain one or more extracted patch regions and 24 have no released extracted patches. The C1 category is under-represented relative to C2 and C5, with 33 extracted patches, and class imbalance should be considered when designing experiments, training models, or interpreting benchmark results. Since patches are variable in size, users may need to resize, crop, or pad patch images before using them in fixed-input-size deep learning pipelines. Patient-level splits should be used for model evaluation to reduce leakage between patches originating from the same patient.
\section*{Data Availability}
The breast FNAC cytology dataset described in this Data Descriptor is available at Zenodo \cite{Patil:2026zenodo}: \url{https://doi.org/10.5281/zenodo.20763900}. The main Zenodo record contains \file{breast-cytology-main.zip}, which includes extracted patch images, WSI-level GeoJSON annotation files, public metadata CSV files, a data dictionary, a metadata README, and \file{zenodo_files_manifest.csv}. The 470 whole-slide images are provided in NDPI format and distributed across 21 linked Zenodo records listed in the main record and in the released manifest. All direct patient identifiers were removed prior to public release.

\section*{Code Availability}
Custom scripts for extracting patches from GeoJSON annotations, parsing GeoJSON annotations, linking extracted patches to parent WSIs, checking metadata consistency, generating patient-level dataset splits, visualizing annotations, and training baseline patch-wise classification models are available at \url{https://github.com/abhijeetptl5/breast-cytology-dataset}. No access restrictions apply.

\section*{Author contributions}
G.J. and A.P. drafted the manuscript under the supervision of S.P. and A.S. G.J., S.P., and A.S. supervised the multi-center dataset development. Clinical and pathology collaborators coordinated sample collection, slide preparation, cytological review, and diagnostic categorization at the participating centers. A.P. organized the digital dataset, metadata files, annotation linkage, Zenodo release, and code repository. All authors reviewed and approved the final manuscript.

\section*{Competing interests}
The authors declare no competing interests.
\section*{Funding} 
Not Applicable
\end{document}